\documentclass[10pt,twocolumn,letterpaper]{article}

\usepackage{iccv}
\usepackage{times}
\usepackage{epsfig}
\usepackage{amsmath,graphicx}
\usepackage{amssymb}
\usepackage{array}
\usepackage{xcolor}
\usepackage{booktabs} 
\usepackage{multirow}
\usepackage{microtype}
\usepackage{bbding}      
\usepackage{booktabs}
\usepackage{pifont}      
\usepackage{fontawesome}
\usepackage{cite}
\usepackage{subcaption}

\usepackage[pagebackref=false,breaklinks=true,letterpaper=true,colorlinks,citecolor=blueish,bookmarks=false]{hyperref}

\usepackage[accsupp]{axessibility}  
\usepackage[referable]{threeparttablex}

\definecolor{brickred}{rgb}{0.8, 0.25, 0.33}
\definecolor{blueish}{rgb}{0.0, 0.3, 0.6}

\usepackage{enumitem}
\setitemize{noitemsep,topsep=0pt,parsep=0pt,partopsep=0pt}

\def\VspaceL{\vspace{-0.00cm}}

\newcolumntype{C}[1]{>{\PreserveBackslash\centering}p{#1}}
\newcolumntype{R}[1]{>{\PreserveBackslash\raggedleft}p{#1}}
\newcolumntype{L}[1]{>{\PreserveBackslash\raggedright}p{#1}}

\makeatletter
\def\thanks#1{\protected@xdef\@thanks{\@thanks
        \protect\footnotetext{#1}}}
\makeatother

\iccvfinalcopy

\def\iccvPaperID{1314}

\ificcvfinal\fi

\begin{document}

\title{Co-Evolution of Pose and Mesh for 3D Human Body Estimation from Video}

\author{
Yingxuan You\textsuperscript{1} \quad
Hong Liu\textsuperscript{1 $\dagger$} \thanks{$\dagger$~Corresponding author.} \quad Ti Wang\textsuperscript{1} \quad Wenhao Li\textsuperscript{1} \quad 
Runwei Ding\textsuperscript{1} \quad Xia Li\textsuperscript{2} \\
\textsuperscript{1}Key Laboratory of Machine Perception, Shenzhen Graduate School, Peking University \\
\textsuperscript{2}Department of Computer Science, ETH Zürich \\
{ 
\tt\small \{youyx,tiwang\}@stu.pku.edu.cn, \{hongliu,wenhaoli,dingrunwei\}@pku.edu.cn, xia.li@inf.ethz.ch}
}

\maketitle

\ificcvfinal\thispagestyle{empty}\fi

\begin{abstract}
Despite significant progress in single image-based 3D human mesh recovery, accurately and smoothly recovering 3D human motion from a video remains challenging. 
Existing video-based methods generally recover human mesh by estimating the complex pose and shape parameters from coupled image features, whose high complexity and low representation ability often result in inconsistent pose motion and limited shape patterns.
To alleviate this issue, we introduce 3D pose as the intermediary and propose a \textbf{P}ose and \textbf{M}esh \textbf{C}o-\textbf{E}volution network~(PMCE) that decouples this task into two parts: 1) video-based 3D human pose estimation and 2) mesh vertices regression from the estimated 3D pose and temporal image feature. Specifically, we propose a two-stream encoder that estimates mid-frame 3D pose and extracts a temporal image feature from the input image sequence. 
In addition, we design a co-evolution decoder that performs pose and mesh interactions with the image-guided Adaptive Layer Normalization~(AdaLN) to make pose and mesh fit the human body shape.
Extensive experiments demonstrate that the proposed PMCE outperforms previous state-of-the-art methods in terms of both per-frame accuracy and temporal consistency on three benchmark datasets: 3DPW, Human3.6M, and MPI-INF-3DHP. Our code is available at \href{https://github.com/kasvii/PMCE}{https://github.com/kasvii/PMCE}.
\end{abstract}

\section{Introduction}
Recovering 3D human mesh from an image or a video is an essential yet challenging task for many applications, such as human-robot interaction, virtual reality, and motion analysis. 
The challenges of this task arise from the 2D-to-3D ambiguity, cluttered background, and occlusions. 
Recently, many studies~\cite{pavlakos2018learning, kanazawa2018end, choi2020pose2mesh, kocabas2021pare, li2021hybrik, li2022cliff} have been proposed to recover the 3D human mesh from a single image, which can generally be categorized into RGB-based methods and pose-based methods. RGB-based methods predict human mesh end-to-end from image pixels, typically predicting the pose and shape parameters of the parametric human model~(\emph{e.g.}, SMPL~\cite{loper2015smpl}) to generate the 3D human mesh. 
However, the representation ability of the parametric model is constrained by the limited pose and shape space~\cite{kolotouros2019convolutional, li2021hybrik}. To overcome this limitation, non-parametric approaches have been proposed to predict the 3D coordinates of mesh vertices directly, which generally use Graph Convolutional Networks~(GCNs)~\cite{choi2020pose2mesh, pqgcn} or Transformers~\cite{cho2022FastMETRO, zheng2021lightweight, lin2021mesh} to capture the relations among vertices. 
In contrast, pose-based methods leverage 2D pose detectors~\cite{chen2018cascaded, sun2019deep} as the front-end to recover human mesh from the detected 2D poses.
With the significant advancements in 2D pose detection, pose-based methods have become increasingly robust and lightweight, making them popular for real-world applications~\cite{zheng2021lightweight}.

\begin{figure}[t]
\centering    
\includegraphics[width=1.02\linewidth]{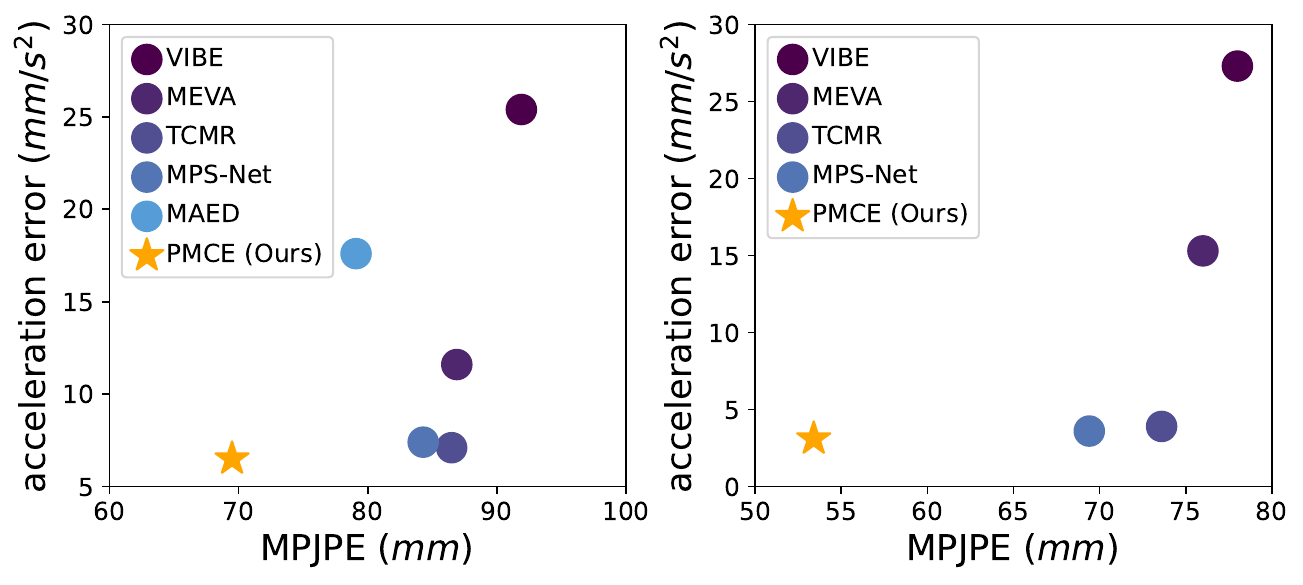}
\caption{Comparison with video-based 3D human mesh recovery methods upon accuracy~(MPJPE~$\downarrow$) and temporal consistency~(acceleration error~$\downarrow$) on 3DPW~\cite{von2018recovering}~(left) and Human3.6M~\cite{ionescu2013human3}~(right) datasets. 
}
\label{fig:topfig}
\VspaceL
\end{figure}

Despite the significant progress in single-image 3D human mesh recovery, these methods still struggle to capture temporally consistent human motion from videos.
To solve this problem, several works~\cite{kocabas2020vibe, choi2021beyond, luo2020,wei2022capturing} extend single image-based methods to video cases.
They use a pre-trained Convolutional Neural Network~(CNN)~\cite{kolotouros2019learning} to extract static features for each frame, then train a temporal network to predict SMPL parameters.
Although these video-based methods significantly improve the temporal consistency of 3D human motion, there exists a trade-off between per-frame accuracy and motion smoothness for the following two main reasons: 
1) The highly coupled image feature. 
The extracted image features in deep CNN layers are low-resolution and tightly coupled~\cite{sun2019human}, which inevitably discard the spatial information in the image~\cite{wan2021encoder}.
2) The limited representation ability of the parametric human model. The SMPL model represents pose using 3D rotation, which might face periodicity and discontinuity issues~\cite{kolotouros2019convolutional}, making pose prediction in videos more challenging.
Besides, the local and swift mesh deformations described by the shape parameters are difficult to learn.

In recent years, video-based 3D human pose estimation from the detected 2D poses has achieved high pose accuracy and motion smoothness~\cite{mhformer, zhang2022mixste}, which inspires us that the skeleton sequence contains sufficient spatial and temporal pose information of human motion. But for the mesh recovery task, detailed shape information is needed, which can be acquired from the image features.
Based on the above observations, 
we propose the \textbf{P}ose and \textbf{M}esh \textbf{C}o-\textbf{E}volution network~(PMCE) to recover 3D human mesh from videos in a non-parametric way. We decouple the 3D human mesh recovery task into two consecutive parts: 1) video-based 3D pose estimation and 2) mesh vertices regression from 3D pose and image feature, where the latter is the focus. Specifically, in the first part, 
we propose a two-stream encoder. One stream takes a 2D pose sequence detected from input images to estimate the mid-frame 3D pose, and the other stream takes static image features extracted from images and aggregates them for a temporal image feature.
In the second part, we design a co-evolution decoder that performs pose and mesh interactions with an image-guided Adaptive Layer Normalization~(AdaLN). AdaLN adjusts the statistical characteristics of joint and vertex features to make the pose and mesh better fit the human body shape. 
As shown in Figure~\ref{fig:topfig}, compared to previous video-based methods, PMCE achieves better performance in terms of per-frame accuracy and temporal consistency on 3DPW~\cite{von2018recovering} and Human3.6M~\cite{ionescu2013human3} datasets.

In summary, our contributions are as follows:
\begin{itemize}
\item We propose a Pose and Mesh Co-Evolution network~(PMCE) for recovering 3D human mesh from video. It decouples the task into two parts: video-based 3D pose estimation, and mesh vertices regression 
by image-guided pose and mesh co-evolution, 
achieving accurate and temporally consistent results.

\item We design the co-evolution decoder that performs pose and mesh interactions guided by our proposed AdaLN. AdaLN adjusts the statistical characteristics of joint and vertex features based on the image feature to make them conform to the human body shape.
\item Our method achieves state-of-the-art performance on challenging datasets like 3DPW, reducing MPJPE by 12.1\%, PVE by 8.4\% and acceleration error by 8.5\%.
\end{itemize}

\begin{figure*}[htb]
    \centering    
    \includegraphics[width=\linewidth]{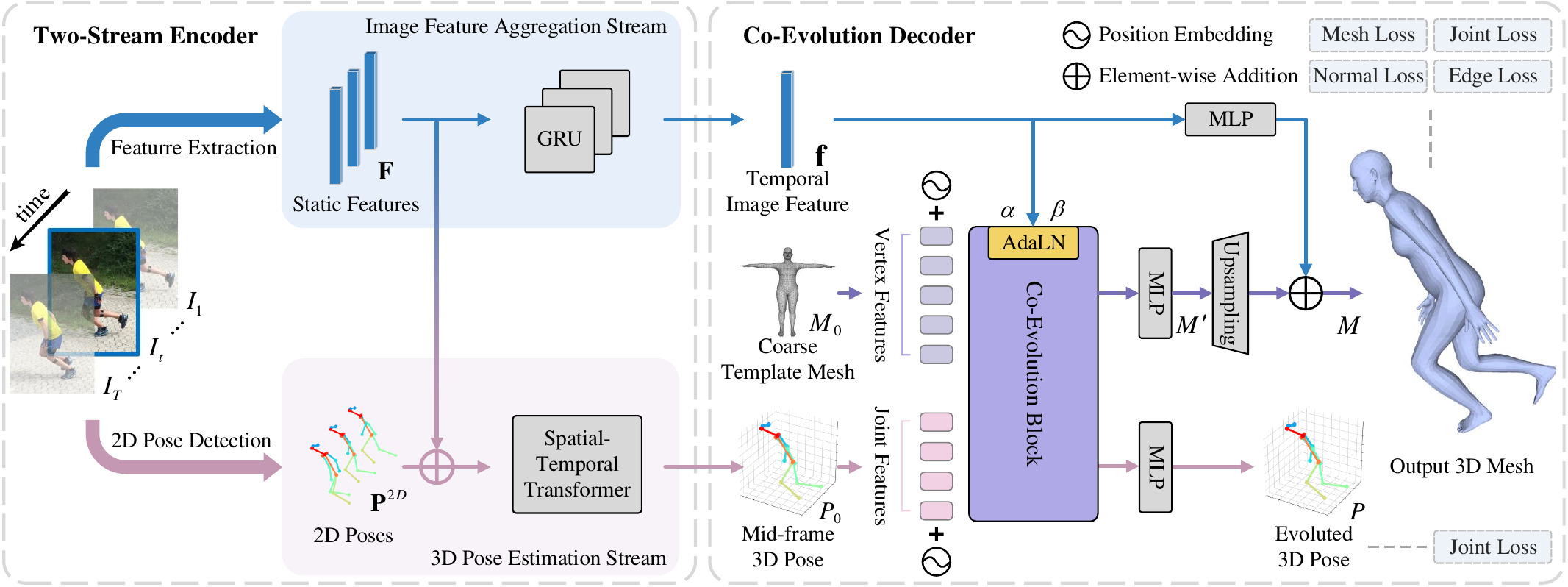}
    \caption{Architecture of the proposed PMCE. Given a video sequence, static image features are extracted by a pre-trained CNN~\cite{kolotouros2019learning}, and 2D poses are detected by an off-the-shelf 2D pose detector~\cite{chen2018cascaded, xu2022vitpose}. The two-stream encoder leverages dual parallel modules to generate a temporal feature and estimate the mid-frame 3D pose, respectively. Then, the co-evolution decoder
        regresses the mesh vertices through the pose and mesh interactions guided by our proposed AdaLN, which makes the pose and mesh fit the body shape.
    }
    \label{fig:archi}
    \VspaceL
\end{figure*}

\section{Related Work}
\noindent \textbf{Estimating 3D human pose from monocular video.}
Benefiting from the advance of 2D pose detection, video-based 3D human pose estimation often takes a 2D pose sequence detected from images by an off-the-shift 2D pose detector to estimate the 3D pose. 
Recent years have witnessed the development of video-based 3D human pose estimation. Several methods have been proposed to explore spatial and temporal pose information through Temporal Convolution Networks~(TCNs)~\cite{pavllo20193d,liu2020attention,zeng2020srnet,chen2021anatomy}, and Transformer-based networks~\cite{zheng20213d,li2023exploiting,mhformer,zhang2022mixste}, resulting in an impressive performance in terms of pose accuracy and motion smoothness.
The achievements in video-based 3D human pose estimation demonstrate that the pose sequence contains sufficient spatial and temporal information, which can be leveraged to produce accurate and smooth 3D human pose motion. 

\noindent \textbf{Recovering 3D human mesh from a single image.}
3D human mesh recovery aims to reconstruct 3D human pose and shape from monocular images. Previous studies can be divided into parametric methods and non-parametric methods. 
Parametric methods adopt the pre-trained parametric human model~(\emph{e.g.}, SMPL~\cite{loper2015smpl}) and estimate the pose and shape parameters to recover the human mesh. Early works~\cite{kanazawa2018end} directly predict the parameter from the input image, but it is challenging to obtain accurate results. Recent works further introduce human body information extracted from the input image, such as 2D pose~\cite{pavlakos2018learning, yu2021skeleton2mesh}, human part segmentation~\cite{omran2018neural, kocabas2021pare}, silhouette~\cite{pavlakos2018learning, yu2021skeleton2mesh}, and depth map
~\cite{guler2018densepose,zhang2020learning,xu2021learning}, then predict the SMPL parameters from them. 
However, parametric methods are constrained by the limited pose and shape representation space. To this end, non-parametric methods are proposed that directly regress the mesh vertices.
They model the relations between mesh vertex by GCNs~\cite{kolotouros2019convolutional, choi2020pose2mesh, pqgcn} or Transformers~\cite{lin2021end, lin2021mesh, cho2022FastMETRO}, which produce flexible human meshes.
Although the single image-based methods have achieved significant performance in accuracy, it is hard for them to produce temporally consistent and smooth 3D human motion when applied to video cases.

\noindent \textbf{Recovering 3D human mesh from monocular video.}
Different from image-based methods, video-based methods aim to produce accurate and temporally consistent human mesh from video frames. 
Most methods~\cite{kanazawa2019learning, kocabas2020vibe, luo2020, choi2021beyond, wei2022capturing} extract the static image feature vectors using a pre-trained CNN~\cite{kanazawa2018end} with global pooling operation and train the temporal networks, such as 1D CNNs~\cite{kanazawa2019learning}, Gated Recurrent Units~(GRUs)~\cite{kocabas2020vibe, luo2020, choi2021beyond}, and Transformers~\cite{wei2022capturing}, to predict the pose and shape parameters for human mesh representation.
For larger receptive fields, some methods~\cite{sun2019human, wan2021encoder} take the image feature maps before the global pooling as input. 
Sun \textit{et al.}~\cite{sun2019human} disentangle the skeletons from the image feature maps, and then aggregates per-frame skeleton and image feature vector by a bilinear transformation. 
MAED~\cite{wan2021encoder} takes the image patch sequence as input and uses a Transformer-based encoder-decoder framework to predict SMPL parameters.
Despite the promising results of video-based methods, 
the low-resolution image features in the deep CNN layers inevitably discard detailed spatial information. 
Besides, their representation abilities are constrained by the limited parametric space, making it difficult to learn the slight and swift human motion in pose and shape domains.
Therefore, existing video-based methods still struggle to recover accurate and temporally consistent 3D human mesh.
In contrast, our method decouples this task into 3D pose estimation and non-parametric mesh regression by pose and mesh co-evolution, achieving better per-frame accuracy and temporal consistency.

\section{Method}
The overall architecture of our proposed Pose and Mesh Co-Evolution network~(PMCE) is depicted in Figure~\ref{fig:archi}, 
which comprises two sequential steps: 1) video-based 3D pose estimation and 2) mesh vertices regression from the 3D pose and temporal image feature.
The former focuses on the human skeleton and predicts accurate and smooth human motion in terms of the pose. The latter leverages visual cues to supplement the information about the human body shape that recovers accurate human mesh and refines the predicted 3D pose, achieving pose and mesh co-evolution. 
Specifically, given an input video sequence $\mathbf{V}=\{\mathbf{I}_t\}^T_{t=1}$ with $T$ frames, static image features $\mathbf{F} \in \mathbb{R}^{T \times 2048}$ are extracted by a pre-trained ResNet-50~\cite{kolotouros2019learning}, while 2D poses $\mathbf{P}^{2D} \in \mathbb{R}^{T \times J \times 2}$ are detected using an off-the-shelf 2D pose detector~\cite{chen2018cascaded,xu2022vitpose}, where $J$ denotes the number of body joints.
The two-stream encoder applies dual parallel modules to generate a temporal image feature and estimate the mid-frame 3D pose, respectively. 
Then, the co-evolution decoder regresses the coordinates of mesh vertices from the 3D pose and temporal feature through pose and mesh interactions. Our proposed Active Layer Normalization~(AdaLN) guides the interactions by adjusting the statistical characteristics of joint and vertex features based on the temporal image feature to make pose and mesh fit the human body shape. We elaborate on each part in the following subsections.

\subsection{Preliminary}
We design our network based on Transformer to model spatial-temporal pose relations and regress 3D human meshes. 
Thus, we first provide an overview of the fundamental components within the Transformer~\cite{vaswani2017attention}, including Multi-head Self-Attention~(MSA), Multi-head Cross-Attention~(MCA), and Layer Normalization~(LN).

\noindent \textbf{MSA.}
The Transformer's input tokens $X \in \mathbb{R}^{n \times d}$ are linearly projected to quires $Q \in \mathbb{R}^{n \times d}$, keys $K \in \mathbb{R}^{n \times d}$ and values $V \in \mathbb{R}^{n \times d}$. The attention is calculated by the scaled dot product, which can be formulated as follows:
\begin{equation}
\operatorname{Attention}(Q, K, V)=\operatorname{Softmax}\left(Q K^T / \sqrt{d}\right) V,
\end{equation}
where $n$ is the sequence length, and $d$ is the dimension.
$Q, K, V$ are then split into $h$ heads, each performing scaled dot-product attention in parallel. The final output is the concatenation of $h$ heads mapped by a linear projection matrix $W \in \mathbb{R}^{d \times d}$, which
can be expressed as:
\begin{equation}
\begin{aligned}
\operatorname{MSA}(Q, K, V) =\operatorname{Concat}\left(H_1, H_2, \ldots, H_h\right) W, \\
H_i =\operatorname{Attention}\left(Q_i, K_i, V_i\right), i \in[1, \ldots, h],
\end{aligned}
\end{equation}

\noindent \textbf{MCA.}
MCA has a similar structure to MSA. The difference is that, in MCA, the query is projected from $X$, while key and value are projected from another representation $Y$.

\noindent \textbf{LN.}
LN~\cite{ba2016layer} normalizes the activities of the neurons in a layer. Given an input token feature $x \in \mathbb{R}^{d}$, LN can be defined as a linear function adjusted by two trainable parameters, \emph{i.e.}, scaling $\alpha \in \mathbb{R}^{d}$ and shifting $\beta \in \mathbb{R}^{d}$:
\begin{equation}
    \operatorname{LN}(x; \alpha, \beta) = \alpha \odot\left(\frac{x-\mu(x)}{\sigma(x)}\right)+\beta, 
\end{equation}
\begin{equation}
    \mu(x) = \frac{1}{d} \sum_{i=1}^{d} x_i, \quad
    \sigma(x) = \sqrt{\frac{1}{d} \sum_{i=1}^{d}\left(x_i-\mu(x)\right)^2}, \\
\end{equation}
where $\odot$ is element-wise multiplication, $\mu$ and $\sigma$ denotes mean and standard deviation taken across the elements of $x$, respectively. $x$ is first normalized by $\mu$ and $\sigma$ then scaled and shifted by $\alpha$ and $\beta$.

\subsection{Two-Stream Encoder}
\noindent \textbf{2D pose normalization by the full image.}
Previous methods~\cite{choi2020pose2mesh, kanazawa2018end, kocabas2021pare, kocabas2020vibe, wei2022capturing} of 3D human mesh recovery typically follow the top-down manner where the region of humans is detected and cropped before being processed individually. This manner is effective in reducing background noise and simplifying feature extraction. However, the cropping operation discards the location information in the full image, which is essential to predict the global rotation in the original camera coordinate system~\cite{li2022cliff}. To compensate for the loss of location information without introducing additional computation, 
we normalize the 2D pose $\textbf{P}^{2D} \in \mathbb{R}^{T \times J \times 2}$ with respect to the full image instead of the cropped region~\cite{choi2020pose2mesh, zheng2021lightweight}.
The 2D pose normalization is reformulated as:
\begin{equation}
    \overline{\mathbf{P}}^{2D} = \frac{2 \times \mathbf{P}^{2D}}{w} - [1, \frac{h}{w}],
\end{equation}
where $w$ and $h$ denote the width and height of the image, respectively. $\overline{\mathbf{P}}^{2D} \in \mathbb{R}^{T \times J \times 2}$ is the normalized 2D pose sequence within the range of -1 to 1.

\noindent \textbf{3D pose estimation stream.}
Our 3D pose estimation stream is built upon the spatial-temporal Transformer~(ST-Transformer)~\cite{zhang2022mixste, wan2021encoder}, which is designed to estimate the mid-frame 3D pose from the 2D pose sequence (the illustration is provided in Sup. Mat. ). 
Given the normalized 2D pose sequence $\overline{\mathbf{P}}^{2D}$, we first project it to the high-dimensional joint feature $\mathbf{X} \in \mathbb{R}^{T \times J \times C_1}$ with the feature dimension $C_1$. Secondly, we project the static image features $\mathbf{F} \in \mathbb{R}^{T \times 2048}$ to $\mathbf{F}' \in \mathbb{R}^{T \times C_1}$, and expand it to $T \times 1 \times C_1$, then add to the joint feature $\mathbf{X}$. To retain the positional information of spatial and temporal domains, we add the spatial embedding and the temporal embedding of the joint sequence to $\mathbf{X}$. The ST-Transformer consists of $L_1$ layers cascaded spatial Transformer and the temporal Transformer. The spatial Transformer aims to explore the spatial information among joints, which calculates the similarities between joint tokens in the same frame. Moreover, to capture the temporal relation among frames, the temporal Transformer reshapes the joint feature $\mathbf{X}$ from $(T \times J \times C_1)$ to $(J \times T \times C_1)$. Thus, the attention matrix is calculated by the similarities between frame tokens of the same joint. Finally, we use a multi-layer perceptron (MLP) to transform the dimension from $C_1$ to 3 and fuse $T$ frames to one to get the mid-frame 3D pose $P_0 \in \mathbb{R}^{J \times 3}$. 

\noindent \textbf{Image feature aggregation stream.}
The image feature aggregation stream aims to aggregate the static image features of $T$ frames to get a mid-frame temporal feature. Given the static image feature sequence $\mathbf{F} \in \mathbb{R}^{T \times 2048}$, we use a bi-directional GRU~\cite{cho2014learning} to capture the temporal information by recurrently updating the features of neighboring frames. After that, GRU yields an updated temporal image feature $\mathbf{f} \in \mathbb{R}^{2048}$ for mid-frame that aggregates the temporal information of all $T$ frames.

\subsection{Co-Evolution Decoder}
The co-evolution decoder is proposed to recover the plausible human mesh in a non-parametric manner using the 3D pose $P_0$, temporal image feature $\mathbf{f}$, and coarse template mesh $M_0$~(provided by SMPL~\cite{loper2015smpl}) as input, as shown in Figure~\ref{fig:archi}.
The mid-frame 3D pose and temporal image feature generated by the two-stream encoder provide complementary information on pose and shape, respectively.
On the one hand, the 3D pose focuses on skeletal motion, which provides more precise and robust pose information than the image feature. On the other hand, the image feature contains visual cues, such as body shape and surface deformation, which are not available in the sparse 3D pose. 
The complementary information on human pose and shape is crucial for accurate 3D human mesh recovery.
Furthermore, the body shape information, serving as a prior, can also refine the estimated 3D pose.  
Motivated by the above observation, we propose a co-evolution block that performs pose and mesh interactions with an Adaptive Layer Normalization (AdaLN). AdaLN adaptively adjusts the statistical characteristics
of joint and vertex features based on the temporal image feature that guides the pose and mesh to fit the human body shape.

\noindent \textbf{Adaptive layer normalization.} Inspired by the style transfer task~\cite{huang2017arbitrary} that swaps style 
from a style image to a content image 
in the feature space by transferring feature statistics, we propose AdaLN to adaptively adjust the features of joints and vertices towards the image feature $\mathbf{f}$. Different from LN~\cite{ba2016layer}, the scaling $\alpha$ and shifting $\beta$ of AdaLN are generated from the image feature. The formulation can be written as:
\begin{equation}
    \operatorname{AdaLN}(x, \mathbf{f}) = \alpha(\mathbf{f}) \odot\left(\frac{x-\mu(x)}{\sigma(x)}\right)+\beta(\mathbf{f}), 
\end{equation}
\begin{equation}
    \alpha(\mathbf{f}) = \operatorname{MLP}_{\alpha}(\mathbf{f}), \quad \beta(\mathbf{f}) = \operatorname{MLP}_{\beta}(\mathbf{f}),
\end{equation}
where $\alpha$ and $\beta$ are linearly transformed from the image feature $\mathbf{f}$ by two MLPs, respectively. 
AdaLN takes the token feature $x$ and image feature $\mathbf{f}$ as inputs and adjusts the mean and standard deviation of $x$ based on $\mathbf{f}$, allowing $x$ adaptively matches to different $\mathbf{f}$.
In this way, the shape information contained in the image feature can be injected into the joint and vertex features, while preserving their spatial structure.

\begin{figure}[t]
    \centering    
    \includegraphics[width=\linewidth]{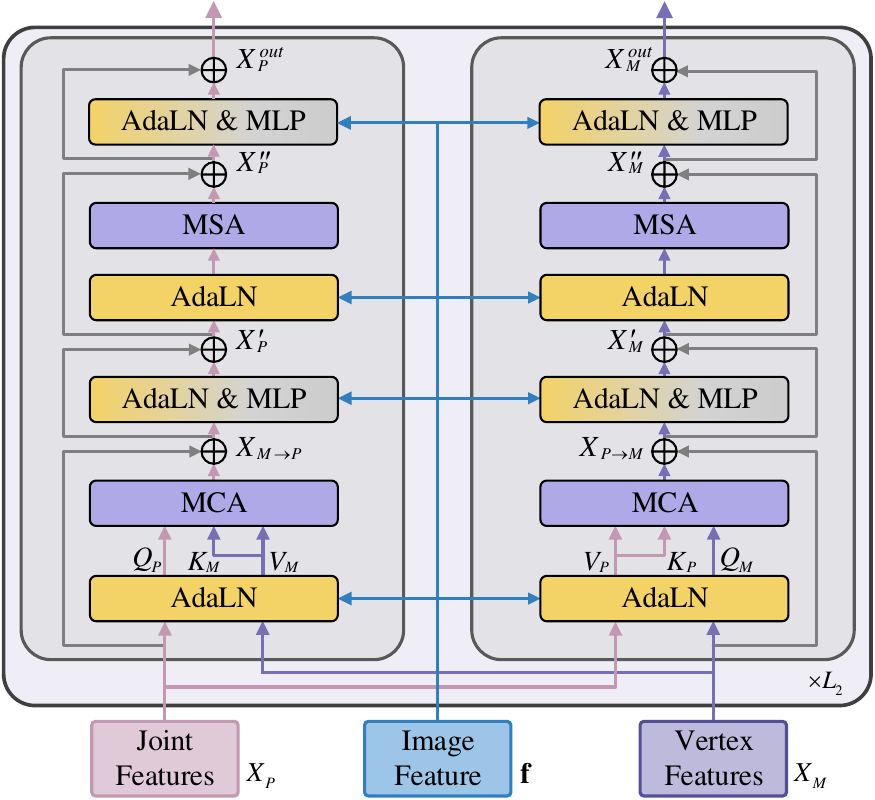}
    \caption{The co-evolution block is designed with a symmetric attention architecture to perform pose and mesh interactions. The joint and vertex features are adjusted by the AdaLN that guides pose and mesh to fit the human body shape based on the image feature.
    }
    \label{fig:adafusion}
    \VspaceL
\end{figure}

\noindent \textbf{Co-evolution block.}
As shown in Figure~\ref{fig:adafusion}, we design the co-evolution block in a symmetric attention mechanism to model the interactions of pose and mesh. 
The inputs are the estimated pose $P_0 \in \mathbb{R}^{J \times 3}$, temporal image feature $\mathbf{f} \in \mathbb{R}^{2048}$, and coarse template mesh $M_0 \in \mathbb{R}^{V' \times 3}$~\cite{loper2015smpl}, where $V'=431$ denotes the vertex number of the coarse mesh. The coordinate of each mesh vertex is reinitialized to that of its nearest 3D joint in $P_0$, where the distances between joints and vertices are calculated from the template 3D pose and mesh~\cite{loper2015smpl}. We first linearly project the pose joints and mesh vertices to high-dimensional features $X_P \in \mathbb{R}^{J \times C_2}$ and $X_M \in \mathbb{R}^{V' \times C_2}$ respectively, then add position embedding to each other. Each feature is normalized by AdaLN with the image feature $\mathbf{f}$.
After that, the vertex feature~(joint feature for another branch) serves as the query $Q$, while joint feature~(vertex feature for another branch) is regarded as key $K$ and value $V$. They are fed to MCA with a residual connection to assist gradient propagation, which can be formulated as:
\begin{equation}
\begin{aligned}
& X_{M \rightarrow P}=\operatorname{MCA}\left(Q_P, K_M, V_M \right) + X_P, \\
& X_{P \rightarrow M}=\operatorname{MCA}\left(Q_M, K_P, V_P \right) + X_M,
\end{aligned}
\end{equation}
where $X_{M \rightarrow P}$ and  $X_{P \rightarrow M}$ are the cross-attention features interacting between joint and vertex representations.
Then, the cross-attention features are merged by an MLP:
\begin{equation}
\begin{aligned}
X_P^{\prime}&=\operatorname{MLP}\left(\operatorname{AdaLN}\left(X_{M \rightarrow P}, \mathbf{f}\right)\right) + X_{M \rightarrow P}, \\
X_M^{\prime}&=\operatorname{MLP}\left(\operatorname{AdaLN}\left(X_{P \rightarrow M}, \mathbf{f}\right)\right) + X_{P \rightarrow M}.
\end{aligned}
\end{equation}

Afterward, we perform self-interaction by MSA and MLP, which can be expressed as:
\begin{equation}
\begin{aligned}
X_{P/M}^{\prime \prime} &= \operatorname{MSA}\left(\operatorname{AdaLN}\left(X_{P/M}^{\prime}, \mathbf{f}\right)\right) + X_{P/M}^{\prime}, \\ 
X_{P/M}^{out}&=\operatorname{MLP}\left(\operatorname{AdaLN}\left(X_{P/M}^{\prime \prime}, \mathbf{f}\right)\right) + X_{P/M}^{\prime \prime},
\end{aligned}
\end{equation}
where $X_{P/M}^{*}$ denotes $X_{P}^{*}$ or $X_{M}^{*}$.
The joint features $X_P^{out} \in \mathbb{R}^{J \times C_2}$ and vertex feature $X_M^{out} \in \mathbb{R}^{V' \times C_2}$ are regressed to the output pose $P \in \mathbb{R}^{J \times 3}$ and coarse mesh $M'\in \mathbb{R}^{V' \times 3}$ by MLPs, respectively. Then, we upsample the coarse mesh $M'$ to the original mesh $M \in \mathbb{R}^{V \times 3}$ by a linear layer, where $V=6890$. Finally, we add the vertex residuals projected from the temporal image feature $\mathbf{f}$ by MLPs to compensate for the mesh details that are lost during upsampling operation, which can be written as:
\begin{equation}
    M = \operatorname{Upsampling}(M') + \operatorname{MLP_2}(\operatorname{MLP_1}(\mathbf{f})^{\operatorname{T}}).
\end{equation}

\begin{table*}[t]
\centering
\footnotesize
\caption{Evaluation of state-of-the-art methods on 3DPW, MPI-INF-3DHP, and Human3.6M datasets. All methods use pre-trained ResNet-50~\cite{kolotouros2019learning} as the backbone to extract static features except MAED~\cite{wan2021encoder}. 
`*' represents using ViT~\cite{dosovitskiy2020vit} as the backbone.
\textbf{Bold}: best; \underline{Underline}: second best.}
\setlength\tabcolsep{1.15mm}
\begin{tabular}{l|cccc|ccc|ccc}
\toprule
\multirow{2}{*}{Method} & \multicolumn{4}{c|}{3DPW} & \multicolumn{3}{c|}{MPI-INF-3DHP} & \multicolumn{3}{c}{Human3.6M} \\
                        & MPJPE~$\downarrow$ & PA-MPJPE~$\downarrow$ & PVE~$\downarrow$ & ACCEL~$\downarrow$ & MPJPE~$\downarrow$ & PA-MPJPE~$\downarrow$ & ACCEL~$\downarrow$ & MPJPE~$\downarrow$ & PA-MPJPE~$\downarrow$ & ACCEL~$\downarrow$ \\
\midrule
HMMR~(CVPR'19)~\cite{kanazawa2019learning} & 116.5 & 72.6 & 139.3 & 15.2 & - & - & - & - & 56.9 & - \\
VIBE~(CVPR'20)~\cite{kocabas2020vibe} & 91.9  & 57.6 & 99.1  & 25.4 & 103.9 & 68.9 & 27.3 & 65.9 & 41.5 & 18.3 \\
MEVA~(ACCV'20)~\cite{luo2020} & 86.9  & 54.7 & -     & 11.6 & 96.4  & 65.4 & 11.1 & 76.0 & 53.2 & 15.3 \\
TCMR~(CVPR'21)~\cite{choi2021beyond} & 86.5  & 52.7 & 102.9 & \underline{7.1}  & 97.6  & 63.5 & \underline{8.5}  & 62.3 & 41.1 & 5.3\\ 
MAED*~(ICCV'21)~\cite{wan2021encoder}    & \underline{79.1} & \textbf{45.7} & \underline{92.6} & 17.6 & \underline{83.6} & \underline{56.2} & - & \underline{56.4} & \underline{38.7}& -\\
MPS-NET~(CVPR'22)~\cite{wei2022capturing} & 84.3 & 52.1 & 99.7 & 7.4 & 96.7  & 62.8 & 9.6  & 69.4 & 47.4 & \underline{3.6} \\
\midrule
PMCE~(Ours)    & \textbf{69.5} & \underline{46.7} & \textbf{84.8} & \textbf{6.5} & \textbf{79.7} & \textbf{54.5} & \textbf{7.1} & \textbf{53.5} & \textbf{37.7}  & \textbf{3.1}\\
\bottomrule
\end{tabular}
\label{tab:compare_sota}
\end{table*}

\subsection{Loss Functions}
Following \cite{choi2020pose2mesh,zheng2021lightweight}, 
the 3D pose estimation stream is trained using the 3D joint loss $\mathcal{L}_{joint}$ to supervise the intermediate 3D pose $P_0$. 
Then the whole network is supervised by four losses: mesh vertex loss $\mathcal{L}_{mesh}$, 3D joint loss $\mathcal{L}_{joint}$, surface normal loss $\mathcal{L}_{normal}$ and surface edge loss $\mathcal{L}_{edge}$. The final loss is calculated as their weighted sum:
\begin{equation}
\mathcal{L}=\lambda_{m} \mathcal{L}_{mesh} +\lambda_{j} \mathcal{L}_{joint} +\lambda_{n} \mathcal{L}_{normal}+\lambda_{e} \mathcal{L}_{edge},
\end{equation}
where $\lambda_{m} {=} 1$, $\lambda_{j} {=} 1$, $\lambda_{n} {=} 0.1$, and $\lambda_{e} {=} 20$.
More details about the loss functions are in the Sup. Mat.

\section{Experiments}
\subsection{Datasets and Evaluation Metrics}
\noindent \textbf{Datasets.} Following previous works~\cite{choi2021beyond,wei2022capturing,wan2021encoder}, we adopt mixed 2D and 3D datasets for training. For 3D datasets, 3DPW~\cite{von2018recovering}, Human3.6M~\cite{ionescu2013human3}, and MPI-INF-3DHP~\cite{mpii3d} contain the annotations of 3D joints and SMPL parameters. For 2D datasets, COCO~\cite{lin2014microsoft} and MPII~\cite{mpii} contain 2D joint annotation with pseudo SMPL parameters from NeuralAnnot~\cite{moon2022neuralannot}. For evaluation, we report the results on Human3.6M, 3DPW, and MPI-INF-3DHP to quantitatively compare with previous methods.

\noindent \textbf{Metrics.} We report results in 
the metrics of per-frame accuracy and temporal consistency.
For accuracy evaluation, we adopt the Mean Per Joint Position Error~(MPJPE), Procrustes-Aligned Mean Per Joint Position Error~(PA-MPJPE), and Per Vertex Error~(PVE), which measure the errors between the estimated results and the ground truth in millimeter~($mm$).
For temporal evaluation, acceleration error~(ACCEL) in $mm/s^2$ is used to report the smoothness of human motion.

\subsection{Implementation Details}
Following previous video-based methods~\cite{kocabas2020vibe, choi2021beyond, wei2022capturing}, we set the sequence length $T$ to $16$ and use ResNet-50 pre-trained in SPIN~\cite{kolotouros2019learning} as the static image feature extractor. For 2D pose detectors, we adopt CPN~\cite{chen2018cascaded} for Human3.6M and ViTPose~\cite{xu2022vitpose} for 3DPW and MPI-INF-3DHP. The whole training is divided into two stages, and both are optimized by Adam. In the first stage, we train the 3D pose estimation stream with a batch size of $64$ and a learning rate of $5 \times 10^{-5}$ for 30 epochs. 
In the second stage, we load the weights of the 3D pose estimation stream and train the whole model end-to-end for $20$ epochs with a batch size of $32$ and a learning rate of $5 \times 10^{-5}$. 
In the 3D pose estimation stream, we use layer number $L_1 = 3$ and feature dimension $C_1 = 256$.
In the co-evolution block, we set layer number $L_2 = 3$ and feature dimension $C_2 =64$.
The network is implemented by PyTorch on one NVIDIA RTX 3090 GPU. 

\begin{figure}[tb]
    \centering    
    \includegraphics[width=\linewidth]{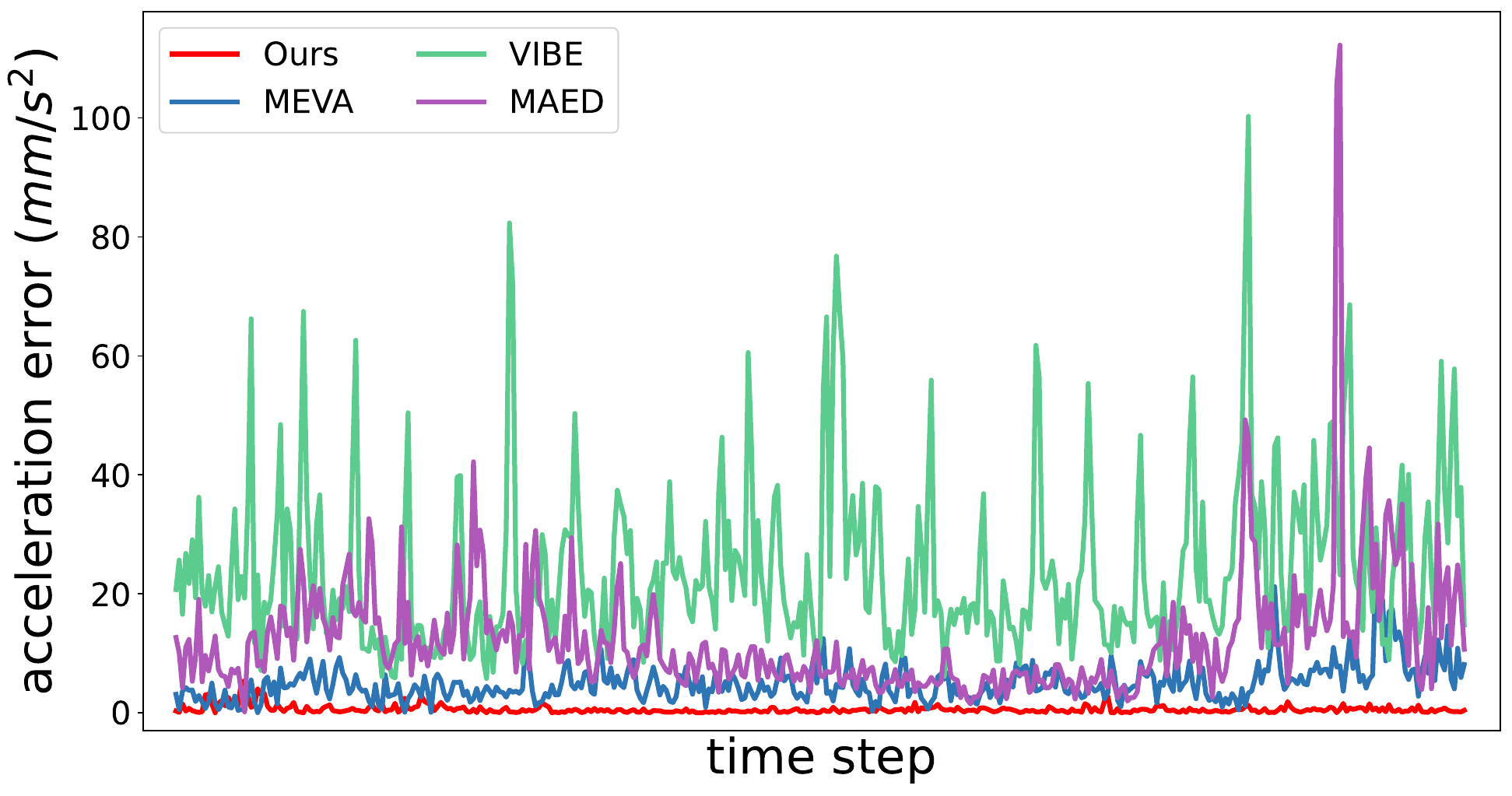}
    \caption{
        Comparison of the acceleration errors for VIBE, MEVA, MAED, and our PMCE.
    }
    \vspace{-1mm}
    \label{fig:acc_err}
    \vspace{-1mm}
\end{figure}

\subsection{Comparison with State-of-the-Art Methods}

\noindent \textbf{Comparison with video-based methods.}
Table~\ref{tab:compare_sota} compares our method with the state-of-the-art video-based methods on the 3DPW, MPI-INF-3DHP, and Human3.6M datasets.
The results show that our method outperforms previous video-based methods, bringing improvements by 12.1\%~(from 79.1 $mm$ to 69.5 $mm$), 4.7\%~(from 83.6 $mm$ to 79.7 $mm$), and 5.1\%~(from 56.4 $mm$ to 53.5 $mm$) on the three datasets in the metric of MPJPE.
Although MAED also makes great progress, it trains ViT~\cite{dosovitskiy2020vit} as the backbone and takes more fine-grained feature maps, which is training time-costing and memory-intensive~\cite{choi2021beyond}. In contrast, other methods~\cite{kocabas2020vibe, choi2021beyond, wei2022capturing} and our PMCE use ResNet-50 pre-trained 
 by SPIN~\cite{kolotouros2019learning} to extract static feature vectors.
Moreover, MAED has a trade-off between per-frame accuracy~(PA-MPJPE) and temporal consistency~(ACCEL). Specifically, when MAED reduces PA-MPJPE by $1$ $mm$, it increases ACCEL by $11.1$ $mm/s^2$ compared to our PMCE on 3DPW.
To further evaluate the temporal consistency, Figure~\ref{fig:acc_err} compares the acceleration errors on the sequence `courtyard\_drinking\_00' of 3DPW. 
Previous methods~\cite{kocabas2020vibe, luo2020, wan2021encoder} reveal large acceleration errors, representing unsmooth and unstable motion estimations. In contrast, our method has relatively low acceleration errors, indicating the temporal consistency of our predictions.
Overall, our method can estimate accurate and smooth 3D human motion from a video. 
The results demonstrate that first using the skeleton sequence to explore spatial-temporal pose information and then regressing the mesh vertices by interacting pose and shape information can effectively achieve per-frame accuracy and temporal consistency. 

\begin{table}[t]
\centering
\footnotesize
\setlength\tabcolsep{0.8mm}
\caption{Comparison with single pose-based methods. All methods are not trained on 3DPW.}
\vspace{-1mm}
\begin{tabular}{l|cccc}
\toprule
\multirow{2}{*}{Method} & \multicolumn{4}{c}{3DPW} \\
                        &  MPJPE~$\downarrow$ & PA-MPJPE~$\downarrow$ & PVE~$\downarrow$ & ACCEL~$\downarrow$ \\
\midrule
PQ-GCN~(TCSVT'22)~\cite{pqgcn} & 89.2 & 58.3 & 106.4& - \\
Pose2Mesh~(ECCV'20)~\cite{choi2020pose2mesh} & 88.9 & 58.3 & 106.3& 22.6 \\
GTRS~(ACM MM'22)~\cite{zheng2021lightweight} & 88.5 & 58.9 & 106.2 & 25.0 \\
\midrule
PMCE~(Ours)     & \textbf{81.6} & \textbf{52.3} & \textbf{99.5} & \textbf{6.8} \\
\bottomrule
\end{tabular}
\label{tab:compare_pose}
\end{table}

\noindent \textbf{Comparison with single pose-based methods.}
We further compare our PMCE to single pose-based methods~\cite{pqgcn, choi2020pose2mesh, zheng2021lightweight} since they also take a detected 2D pose from the image to estimate the intermediate 3D pose and then regress the mesh vertices from the estimated 3D pose, which is relevant to our method. The difference is that our PMCE expands the single pose-based framework to multiple frames and complements it with image cues for 3D human motion estimation.
Table~\ref{tab:compare_pose} compares their performance on the 3DPW dataset. Notably, the pose-based methods did not use the 3DPW training set during training, so all methods are not trained on 3DPW. Our method outperforms pose-based methods in per-frame accuracy and temporal consistency by a large margin, improving by 7.8\% in MPJPE and 69.9\% in ACCEL.

\begin{table}[t]
\centering
\footnotesize
\setlength\tabcolsep{0.35mm}
\caption{Ablation study for the designs of 3D pose estimation stream on Human3.6M. `Seq.' means sequence input.}
\vspace{-1mm}
\begin{tabular}{l|cccc}
\toprule
 Method & MPJPE~$\downarrow$ & PA-MPJPE~$\downarrow$ & PVE~$\downarrow$ & ACCEL~$\downarrow$ \\  
\midrule
Cropped pose Seq.           & 58.2  & 40.8  & 66.3 & 3.3\\
Cropped pose Seq. + bbox info  & 56.4  & 38.6  & 65.0 & 3.4 \\ 
Full image pose Seq.       & \textbf{53.5}  & \textbf{37.7}  & \textbf{61.3} & \textbf{3.1} \\ 
Mid-frame full image pose       & 55.8  & 39.1  & 64.3 & 17.9 \\ 
\bottomrule
\end{tabular}
\label{tab:pose_input}
\vspace{-1mm}
\end{table}

\subsection{Ablation Study}
\noindent \textbf{Effectiveness of 3D pose estimation stream.}
There are two designs in the 3D pose estimation stream, \emph{i.e.} 2D pose normalization by the full image and sequence input, which are respectively evaluated in Table~\ref{tab:pose_input}.
When normalizing the 2D pose in the cropped bounding box, the accuracy is low since the keypoints are located in regions with unfixed locations and scales.
Then, adding bounding box information improves the performance, but the relations between keypoints and bounding boxes are still implicit and hard to explore.
Furthermore, normalizing the 2D pose in full image achieves the best performance since the poses in different frames share the unified coordinate system, which provides clear location information.
Besides, it can be observed that single-frame input~(last line) yields low temporal consistency, while sequence inputs~(top three lines) improve it significantly. 
This result suggests that the 3D pose estimation stream can effectively capture the temporal information of sequence and improves the temporal consistency. 

\begin{table}[t]
\centering
\footnotesize
\caption{Ablation study of image features adding in different modules on 3DPW. `Enc.' denotes the two-stream encoder. 
`Dec.' denotes the co-evolution decoder.
}
\vspace{-1mm}
\setlength\tabcolsep{2.7mm}
\begin{tabular}{l|c|cc}
\toprule
\multirow{2}{*}{Method} & \multicolumn{1}{c|}{Intermediate Pose $P_0$} & \multicolumn{2}{c}{Output Mesh $M$} \\
                        & MPJPE~$\downarrow$           & MPJPE~$\downarrow$ & PVE~$\downarrow$ \\
\midrule
Only pose                   & 72.6 & 76.9 & 92.1  \\ 
$\textbf{F}$ in Enc.        & 70.5 & 72.2 & 89.4  \\  
$\textbf{F}$ in Dec.        & 72.6 & 71.6 & 86.5  \\ 
$\textbf{F}$ in Enc. \& Dec. & \textbf{70.5} & \textbf{69.5} & \textbf{84.8}  \\
\bottomrule
\end{tabular}
\label{tab:imgfeat}
\end{table}

\begin{table}[tb]
\centering
\footnotesize
\setlength\tabcolsep{3.2mm}
\caption{Ablation study for the designs of the co-evolution decoder on 3DPW. 
We report the results of output 3D pose $P$ and output mesh $M$. MPJPE of the input 3D pose is 70.5 $mm$.}
\vspace{-1mm}
\begin{tabular}{l|c|cc}
\toprule
\multirow{2}{*}{Method} & \multicolumn{1}{c|}{Output Pose $P$} & \multicolumn{2}{c}{Output Mesh $M$} \\
                       & MPJPE~$\downarrow$     & MPJPE~$\downarrow$ & PVE~$\downarrow$ \\
\midrule
w/o interactions  & 70.6 & 72.3 & 87.7 \\
Pose$\to$Mesh & 69.7 & 70.8 & 85.7 \\
Mesh$\to$Pose & 68.8 & 71.6 & 87.6 \\
Mesh$\leftrightarrow$Pose~(Ours) &\textbf{67.3} & \textbf{69.5}   & \textbf{84.8} \\ 
\midrule
ALADIN~\cite{ruta2021aladin} & 71.3 & 72.8 & 88.6 \\ 
LN                           & 71.2 & 72.0   & 87.9 \\ 
Bilinear~\cite{sun2019human} & 70.5 & 71.3   & 87.1 \\ 
AdaIN~\cite{huang2017arbitrary} & 70.6 & 71.3 & 86.5 \\ 
AdaNorm~\cite{xu2019understanding} & 69.4 & 70.7 & 86.9 \\
SEAN~\cite{zhu2020sean} & 68.7 & 70.5 & 85.9 \\
AdaLN~(Ours)                 & \textbf{67.3} & \textbf{69.5}   & \textbf{84.8} \\ 
\bottomrule
\end{tabular}
\label{tab:fusion}
\vspace{-1mm}
\end{table}

\noindent \textbf{Effectiveness of image feature.}
Table~\ref{tab:imgfeat} evaluates the impact of image features $\textbf{F}$ on the intermediate 3D pose $P_0$ and the output mesh $M$. When only taking the 2D pose sequence as input, the mesh performance is low since the pose lacks shape information. 
Adding the image feature in the encoder or decoder can improve their individual output results, 
while using the image features in both the encoder and decoder achieves the best results, which indicates that the shape information contained in image features contributes to better results of intermediate 3D pose and output mesh.

\noindent \textbf{Effectiveness of pose and mesh interactions.}
The co-evolution decoder is proposed to regress the mesh vertices from the estimated 3D pose and temporal image feature, containing pose and mesh interactions and AdaLN. 
In the first part of Table~\ref{tab:fusion}, it can be seen that the pose$\to$mesh interaction improves the performance of output mesh $M$ since the mesh can learn more accurate position information guided by the 3D pose. Meanwhile, the mesh$\to$pose interaction enhances the performance of output 3D pose $P$,
which indicates that the priors~(\emph{e.g.}, body shape, limb proportion) in mesh contributes to 3D pose refinement. Moreover, using bi-directional interactions further improves the performance of pose and mesh, which achieves their co-evolution.

\noindent \textbf{Effectiveness of AdaLN.}
In the second part of Table~\ref{tab:fusion}, we evaluate the effect of AdaLN by degenerating it to LN or replacing it with other feature fusion ways, including bilinear fusion~\cite{sun2019human} and some other normalization blocks~\cite{zhu2020sean, ruta2021aladin, huang2017arbitrary, xu2019understanding}. Specifically, bilinear fusion directly combines the image feature with joint and vertex features, which increases the complexity to learn the pose and shape information. The other normalization blocks focus on the fields of 2D image~\cite{zhu2020sean, ruta2021aladin, huang2017arbitrary} or language~\cite{xu2019understanding}, and design normalization approaches in different feature dimensions and statistical parameters towards their specific tasks, but not suitable for the task of 3D human body recovery.
In contrast, AdaLN lies in its tailored adaptation for fitting pose and mesh to individual body shapes by adjusting the statistics of joint and vertex features towards the image features, which enhances the ability of normalization block in 3D human body recovery. 

\begin{figure}[tb]
\centering 
    \begin{subfigure}[b]{0.41\textwidth}
        \centering
        \includegraphics[width=\textwidth]{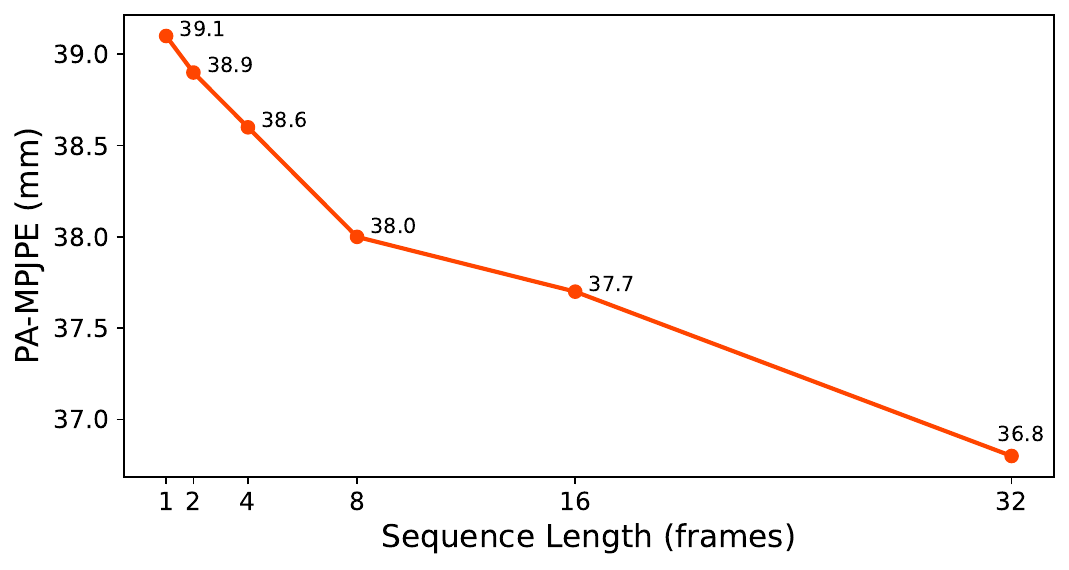}
        \vspace{-5.5mm}
        \caption{Different sequence lengths under PA-MPJPE.}
        \label{fig:subfig1}
        \vspace{2.5mm}
    \end{subfigure}
    \hspace{0.1\textwidth}
    \begin{subfigure}[b]{0.40\textwidth}
        \centering
        \includegraphics[width=\textwidth]{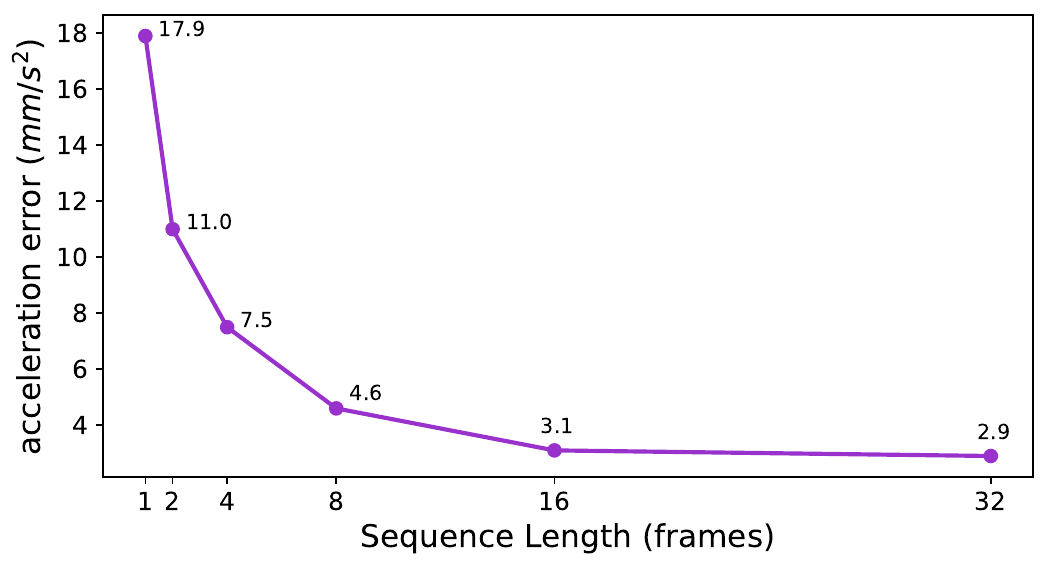}
        \vspace{-5.5mm}
        \caption{Different sequence lengths under acceleration error.}
        \label{fig:subfig2}
    \end{subfigure}
\caption{Ablation studies on different sequence lengths in terms of PA-MPJPE and acceleration error on Human3.6M.}
\label{fig:seqlen}
\vspace{-2mm}
\end{figure}

\noindent \textbf{Impact of sequence lengths.}
For video-based methods, 
sequence length has a direct impact on the results.
Figure~\ref{fig:seqlen} illustrates the influences of sequence length in (a)~accuracy and (b)~temporal consistency on the Human3.6M dataset.  
Increasing the sequence length can improve the performance in terms of both accuracy and temporal consistency. These results indicate that our method can effectively exploit spatial and temporal relations in sequence frames to estimate accurate and smooth 3D human motion. Note that, for a fair comparison with previous video-based methods~\cite{kocabas2020vibe, choi2021beyond, wan2021encoder}, we choose the sequence length as 16 in our experiments.

\noindent \textbf{Impact of 2D pose detections.}
2D poses are used in the 3D pose estimation stream. To evaluate their impact, we conduct experiments in different 2D poses, including the ground truth 2D poses and some detected 2D poses~\cite{chen2018cascaded, pavllo20193d, newell2016stacked}.
Table~\ref{tab:2d_pose} illustrates the results on Human3.6M. Our method is robust to different 2D pose detections, and adopting the 2D pose detectors with higher precision can improve our reconstructed accuracy. The performance with ground truth 2D poses indicates the lower bound of the proposed PMCE, which shows the potential of our method to stay effective with the development of 2D pose detectors.

\begin{table}[t]
\centering
\footnotesize
\setlength\tabcolsep{0.8mm}
\caption{Ablation study of different 2D pose detections on Human3.6M.}
\vspace{-1mm}
\begin{tabular}{l|cccc}
\toprule
{2D pose detection} &  MPJPE~$\downarrow$ & PA-MPJPE~$\downarrow$ & PVE~$\downarrow$ & ACCEL~$\downarrow$ \\
\midrule
Detected by SH~\cite{newell2016stacked} & 56.4 & 39.0 & 64.5 & 3.2 \\
Detected by Detectron~\cite{pavllo20193d} & 55.9 & 39.0 & 64.1 & 3.2 \\
Detected by CPN~\cite{chen2018cascaded} & 53.5 & 37.7 & 61.3 & 3.1 \\
Ground truth 2D pose & \textbf{36.3} & \textbf{26.8} & \textbf{46.2} &  \textbf{2.2} \\
\bottomrule
\end{tabular}
\label{tab:2d_pose}
\vspace{-1mm}
\end{table}

\begin{figure*}[tb]
\centering    
\includegraphics[width=\linewidth]{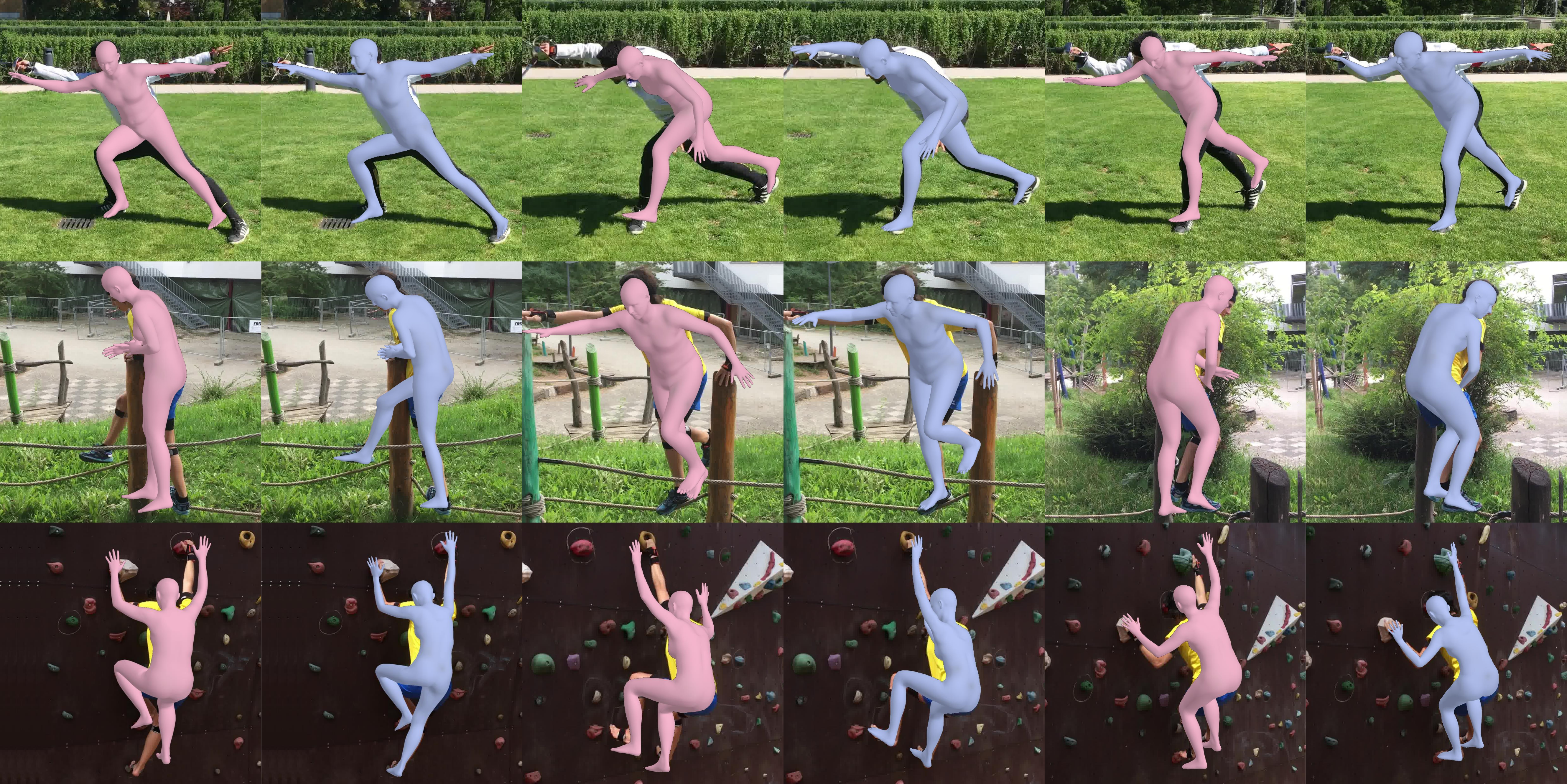}
\caption{Visual comparison between MPS-Net~\cite{wei2022capturing}~(pink meshes) and our PMCE~(blue meshes) on the challenging 3DPW dataset, which contains hard poses, fast motions, and occlusions. Our method can generate more plausible mesh results than MPS-Net.
}
\label{fig:com_mps}
\end{figure*}

\begin{figure*}[tb]
\centering    
\includegraphics[width=\linewidth]{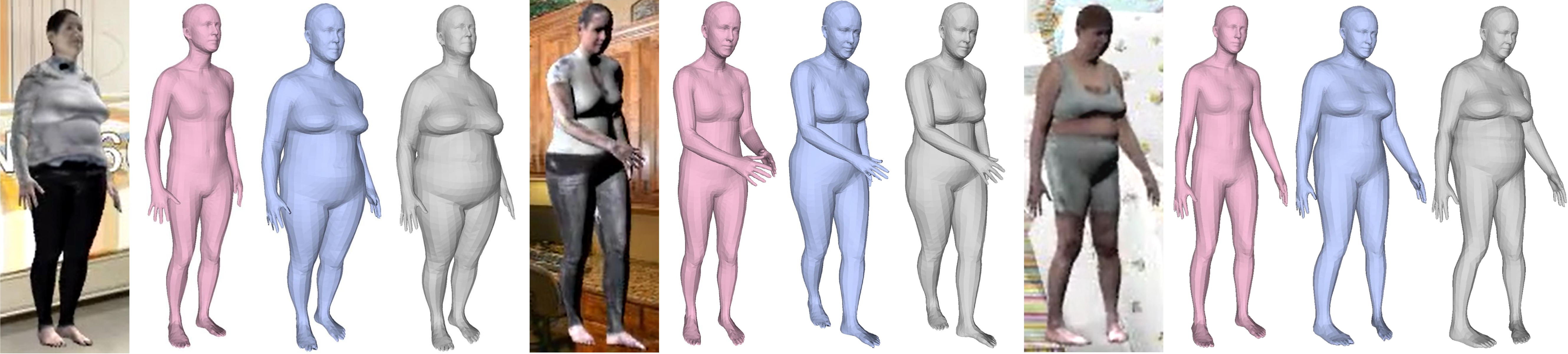}
\caption{Mesh results in extreme shapes. Pink meshes are from MPS-Net~\cite{wei2022capturing}, blue meshes are from PMCE, and grey meshes are ground truth. Our PMCE can fit the body shapes better.
}
\label{fig:shape_comp}
\end{figure*}

\subsection{Qualitative evaluation}
\noindent \textbf{Visual comparison with MPS-Net.}
Figure~\ref{fig:com_mps} shows the qualitative comparison between the previous state-of-the-art method MPS-Net~\cite{wei2022capturing} and our PMCE on the in-the-wild 3DPW dataset. Our method can generate more plausible human meshes, especially on the arms and legs. When an occlusion occurs~(the first sample in Row 2), our PMCE can infer the accurate result based on spatial and temporal relations in a video sequence.

\noindent \textbf{Visualization of human body shape.}
Previous video-based methods~\cite{kocabas2020vibe, choi2021beyond, wei2022capturing} take 10-dimensional shape parameters of SMPL to control human body shape and supervise it in the range of mean shape, which limits their representation ability for various body shapes. On the contrary, our method relieves the limitation by directly regressing mesh vertices and fitting the mesh to the body shape through the proposed AdaLN. 
To verify the representation ability, we train and test MPS-Net~\cite{wei2022capturing} and our PMCE on SURREAL~\cite{varol2017learning} dataset, which contains various body shapes. 
Figure~\ref{fig:shape_comp} shows that MPS-Net tends to produce mean shapes, which reflect unreal body patterns. In contrast, the meshes generated by PMCE are more flexible and align better with the input images. It demonstrates that our PMCE has the ability to learn different human body shapes and recover plausible 3D human meshes. 

\begin{figure}[htbp]
		\centering
		\includegraphics[width=0.82\linewidth]{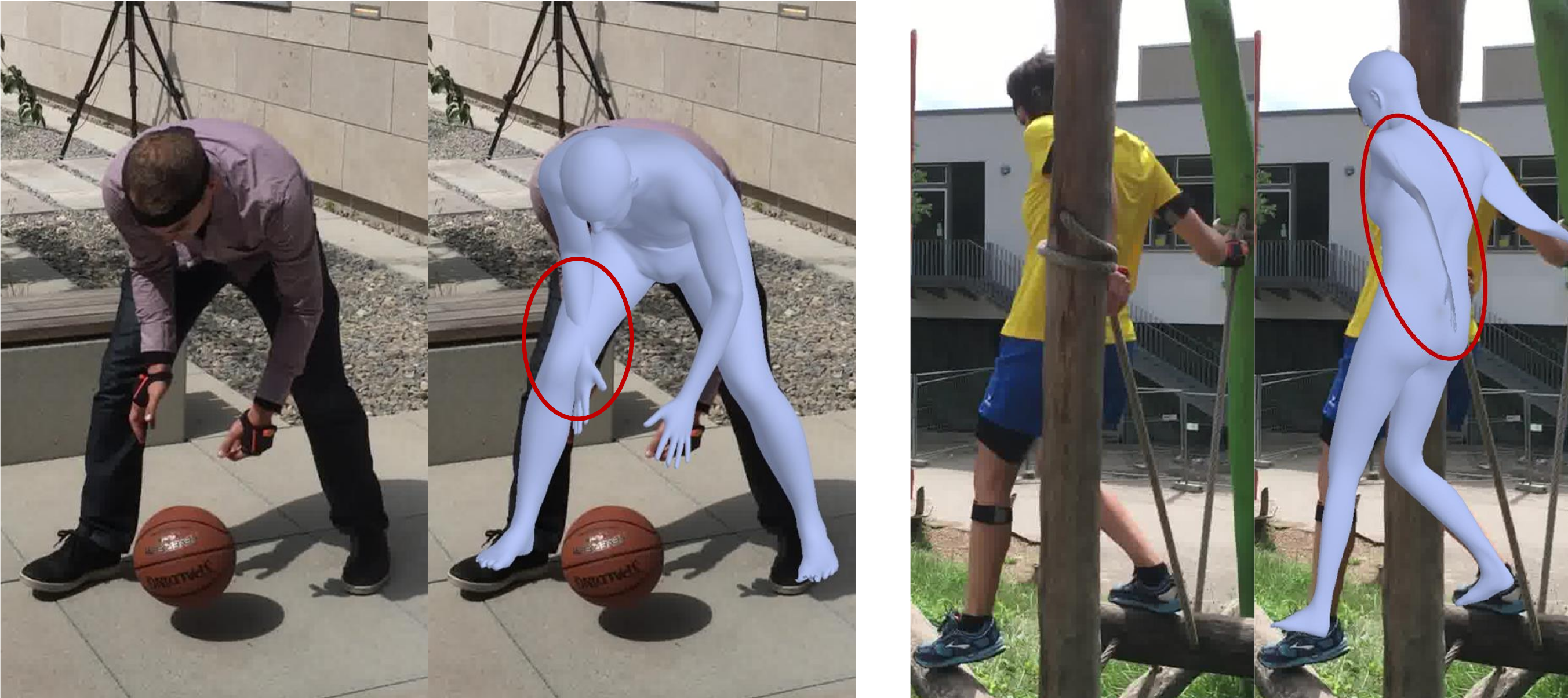}
		\caption{Failure cases in self-contact and challenging pose.}
		\label{chutian2}
\vspace{-3mm}
\end{figure}

\section{Conclusion}
This paper proposes the Pose and Mesh Co-Evolution network~(PMCE), a new two-stage pose-to-mesh framework for recovering 3D human mesh from a monocular video. PMCE first estimates 3D human pose motion in terms of spatial and temporal domains, then performs image-guided pose and mesh interactions by our proposed AdaLN that injects body shape information while preserving their spatial structure. Extensive experiments on popular datasets show that PMCE outperforms state-of-the-art methods in both per-frame accuracy and temporal consistency. We hope that our approach will spark further research in 3D human motion estimation considering both pose and shape consistency.

\noindent \textbf{Limitation.}~Main limitation of our method comes from self-contacts and challenging poses, as shown in Figure~\ref{chutian2}. Due to the low contact perception and limited human body priors, the reconstructed meshes may be implausible. 
Our future work will explore the physical constraints of human body to alleviate this issue.

\noindent \textbf{Acknowledgment.} This work is supported by the National Natural Science Foundation of China (No. 62073004), Basic and Applied Basic Research Foundation of Guangdong (No. 2020A1515110370), Shenzhen Fundamental Research Program (GXWD20201231165807007-20200807164903001, No. JCYJ20200109140410340), the interdisciplinary doctoral grants (iDoc 2021-360) from the Personalized Health and Related Technologies (PHRT) of the ETH domain. 

{\small
\bibliographystyle{ieee_fullname}
\bibliography{ref}
}

\clearpage

\noindent \textbf{\Large{Supplemental Material}}
\vspace{4mm}

\begin{figure*}[h]
\centering    
\includegraphics[width=\linewidth]{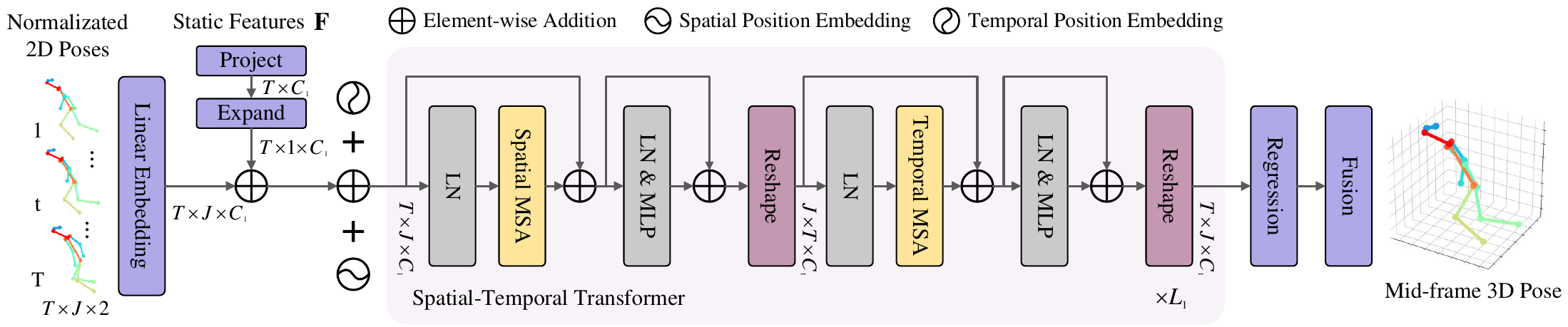}
\vspace{-2mm}
\caption{Architecture of 3D Pose Estimation Stream.}
\label{fig:pose_arch}
\vspace{-1mm}
\end{figure*}

\noindent This supplemental material contains the following parts:

(A)~The architecture of 3D pose estimation stream. 

(B)~Additional quantitative results. 

(C)~Additional ablation study. 

(D)~Details about loss functions. 

(E)~Additional visualization results.

\vspace{4mm}
\noindent \textbf{\large{A. Architecture of 3D Pose Estimation Stream}}
\vspace{2mm}

Figure~\ref{fig:pose_arch} shows the detailed architecture of the 3D pose estimation stream. 
Firstly, the normalized 2D pose sequence is projected to high-dimensional joint features by a linear embedding layer. 
Secondly, we project and expand the static image features, which are added to their corresponding joint features in the same frame. Then we add the spatial and temporal embeddings to joint features and feed joint features to the spatial-temporal Transformer, which consists of cascaded spatial and temporal parts. In the spatial part, the spatial MSA calculates the similarities between joint tokens in the same frame. In the temporal part, the joint features
are reshaped from $(T \times J \times C_1)$ to $(J \times T \times C_1)$, and thus the temporal MSA can calculate the similarities between frame tokens of the same joint. Finally, the joint features are regressed from $C_1$ to $3$ and fused from $T$ frames to one frame to get the mid-frame 3D pose.

\vspace{4mm}
\noindent \textbf{\large{B. Additional Quantitative Results}}
\vspace{2mm}

\noindent \textbf{Comparison with Single RGB-Based Methods.}
Table~\ref{tab:compare_img} compares our PMCE with single RGB-based methods on the 3DPW dataset. All methods use ResNet as the backbone. We evaluate the models trained with and without 3DPW training set for fair comparisons. Single RGB-based methods focus on per-frame accuracy and propose advanced networks to extract image features~\cite{moon2020i2l, zhang2021pymaf, kocabas2021pare, cho2022FastMETRO} and generate human mesh, which shows high performance. In contrast, our PMCE takes pre-trained backbone~\cite{kolotouros2019learning} to extract feature vectors following previous video-based methods~\cite{kocabas2020vibe, choi2021beyond, wei2022capturing}. Compared to the single RGB-based methods, our PMCE achieves competitive performance in PA-MPJPE and outperforms the state-of-the-art method in the metrics of MPJPE, PVE, and ACCEL. The results demonstrate the superiority and effectiveness of our pose and mesh co-evolution design in terms of both per-frame accuracy and temporal consistency for 3D human motion estimation.

\begin{table}[t]
\centering
\footnotesize
\setlength\tabcolsep{1.6mm}
\caption{Comparison with single RGB-based methods. All methods use ResNet as the backbone. `$\dagger$' represents training w/o 3DPW training dataset. `$*$' represents training with 3DPW training set. The top two best results are highlighted in bold and underlined, respectively.}
\vspace{-1mm}
\scalebox{0.95}{
\begin{tabular}{ll|cccc}
\toprule
 & \multirow{2}{*}{Method} & \multicolumn{4}{c}{3DPW} \\
                        & & MPJPE~$\downarrow$ & PA-MPJPE~$\downarrow$ & PVE~$\downarrow$ & ACCEL~$\downarrow$ \\
\midrule
\multirow{10}{*}{\rotatebox{90}{\scriptsize{RGB-based}}}
& HMR$\dagger$~\cite{kanazawa2018end}         & 130   & 76.7 & - & 37.4 \\
& GraphCMR$\dagger$~\cite{kolotouros2019convolutional}    & -     & 70.2 &-&- \\
& SPIN$\dagger$~\cite{kolotouros2019learning}        & 96.9  & 59.2 & 116.4  & 29.8 \\
& I2L-MeshNet$\dagger$~\cite{moon2020i2l} & 93.2  & 57.7 & 110.1  & 30.9 \\ 
& PyMAF$\dagger$~\cite{zhang2021pymaf}       & 92.8  & 58.9 & 110.1  & - \\
& PARE$\dagger$~\cite{kocabas2021pare}        & 82.9  & 52.3 & 99.7   & - \\
& ROMP$*$~\cite{sun2021monocular}              & 79.7  & 49.7 & 94.7  & - \\
& METRO$*$~\cite{cho2022FastMETRO}             & 77.1  & 47.9 & 88.2   & - \\
& CLIFF$*$~\cite{li2022cliff}             & \underline{72.0}  & \textbf{45.7} & \underline{85.3}   & 24.7 \\
\midrule
& PMCE~(Ours)$\dagger$     & 81.6 & 52.3 & 99.5 & \underline{6.8} \\
& PMCE~(Ours)$*$    & \textbf{69.5} & \underline{46.7} & \textbf{84.8} & \textbf{6.5} \\
\bottomrule
\end{tabular}
}
\label{tab:compare_img}
\vspace{1mm}
\end{table}

\noindent \textbf{Generalization in Unseen Views.}
Our method decouples 2D poses and image features from image sequences, which can not only provide complementary pose and shape information for better mesh estimation but also improve the generalization. To verify the latter, we compare our PMCE with the only-pose model~(PMCE without using the image features) on the Human3.6M dataset. Specifically, based on the four camera views of Human3.6M, we train the networks on View 1, View 2, and View 3, then test them on the unseen View 4 to evaluate their generalization in unseen views. As shown in Table~\ref{tab:unseen_view}, the only-pose model suffers from the domain gap between training and testing views, especially when training with few view data~(top line). In contrast, our PMCE has better generalization ability and improves performance by a large margin. The results indicate that our method, complementing the pose information and image features, is effective for a robust mesh estimation in unseen views.

\begin{table}[t]
\centering
\footnotesize
\setlength\tabcolsep{1.15mm}
\caption{Generalization evaluation in unseen views on Human3.6M. The test view is View~4.}
\vspace{-1mm}
\begin{tabular}{l|cc|cc|cc}
\toprule
\multirow{2}{*}{Training views} & \multicolumn{2}{c|}{Only-pose model} & \multicolumn{2}{c|}{PMCE} & \multicolumn{2}{c}{Improvements}\\
                        & MPJPE~$\downarrow$ & PVE~$\downarrow$ & MPJPE~$\downarrow$ & PVE~$\downarrow$ & MPJPE & PVE\\
\midrule
1 & 161.7 & 165.3 & 82.9 & 89.4 & 78.8 & 75.9\\
1, 2 & 100.2 & 112.7 & 59.2 & 69.9 & 40.9 & 42.8 \\
1, 2, 3 & 85.8 & 96.0 & 58.4 & 67.1 & 27.4 & 28.9\\
\bottomrule
\end{tabular}
\label{tab:unseen_view}
\vspace{1mm}
\end{table}

\vspace{4mm}
\noindent \textbf{\large{C. Additional Ablation Study}}
\vspace{2mm}

\begin{table}[t]
\centering
\footnotesize
\caption{Performance comparison between different initializations of mesh vertices on 3DPW.}
\vspace{-1mm}
\setlength\tabcolsep{3.2mm}
\begin{tabular}{l|cccc}
\toprule
Mesh initialization & MPJPE~$\downarrow$ & PA-MPJPE~$\downarrow$ & PVE~$\downarrow$ \\ 
\midrule
Zeros                       & 72.3     & 48.5    & 88.5 \\ 
Template                          & 71.4     & 47.6    & 86.2 \\ 
Nearest joints~(Ours) & \textbf{69.5} & \textbf{46.7}  & \textbf{84.8}   \\
\bottomrule
\end{tabular}
\label{tab:init}
\end{table}

\noindent \textbf{Impact of Mesh Initializations.}
Mesh initialization serves as a human body prior in our method. Table~\ref{tab:init} examines the impact of different mesh initializations, including setting mesh vertices to zeros, using T-shape template mesh from SMPL~\cite{loper2015smpl} or setting the position of per mesh vertex as that of its nearest joint in estimated 3D pose $P_0$~(the distances between vertices and joints are pre-calculated from the template mesh and pose provided by SMPL). Compared with template mesh, setting vertices to their nearest joints makes the initialized mesh closer to the final mesh, which can provide a more precise human body prior and contribute to the final mesh performance.

\vspace{4mm}
\noindent \textbf{\large{D. Loss Functions}}
\vspace{2mm}

For the 3D pose estimation stream, we use the L1 joint loss to supervise the intermediate 3D pose $P_0$, which is defined as follows:
\begin{equation}
\mathcal{L}_{joint}^{int}=\frac{1}{J} \sum_{i=1}^{J}\left\|P_{gt}-{P_0}\right\|_{1}.
\end{equation}

After training the 3D pose estimation stream, we train the whole network using the following four loss functions.

\noindent \textbf{Mesh Loss.} We use the L1 loss between the ground truth 3D mesh vertices $M_{gt} {\in} \mathbb{R}^{V \times 3}$ and the predicted 3D mesh vertices $M {\in} \mathbb{R}^{ V \times 3}$. The mesh vertex loss is calculated as:
\begin{equation}
\mathcal{L}_{mesh}=\frac{1}{V} \sum_{i=1}^{V}\left\|M_{gt}-M\right\|_{1}.
\end{equation}

\noindent \textbf{Joint Loss.} We multiply the predicted 3D mesh $M$ by a predefined matrix $\mathcal{J} {\in} \mathbb{R}^{J \times V}$ to obtain the regressed 3D joints and calculate the joint loss with ground truth 3D joints $P_{gt}$:
\begin{equation}
\mathcal{L}_{joint}=\frac{1}{J} \sum_{i=1}^{J}\left\|P_{gt}-\mathcal{J}M\right\|_{1}.
\end{equation}

\noindent \textbf{Surface Normal Loss.} This loss is used to improve surface smoothness and local details. It is calculated by the normal vectors of the ground truth mesh and the predicted mesh:
\begin{equation}
\mathcal{L}_{normal}=\sum_{f} \sum_{\{i, j\} \subset f}\left|\left\langle\frac{m_{i}-m_{j}}{\left\|m_{i}-m_{j}\right\|_{2}}, n_{gt}\right\rangle\right|,
\end{equation}
where $f$ denotes a triangle face in the mesh, $m_{i}$ and $m_{j}$ denote the $i_{th}$ and $j_{th}$ mesh vertices of the triangle face respectively. And $n_{gt}$ is the unit normal vector of the triangle face $f$ in the ground truth mesh.

\noindent \textbf{Surface Edge Loss.} This loss is used to improve the smoothness of the areas with dense vertices, \emph{e.g.}, hands and feet. The edge length consistency loss is calculated by the ground truth edges and the predicted edges as:
\begin{equation}
\mathcal{L}_{edge}=\sum_{f} \sum_{\{i, j\} \subset f}\left|\left\|m_{gt_i}-m_{gt_j}\right\|_{2}-\left\|m_{i}-m_{j}\right\|_{2}\right|.
\end{equation}

Given the four loss functions, the final loss is calculated as the weighted sum:
\begin{equation}
\mathcal{L}=\lambda_{m} \mathcal{L}_{mesh} +\lambda_{j} \mathcal{L}_{joint} +\lambda_{n} \mathcal{L}_{normal}+\lambda_{e} \mathcal{L}_{edge},
\end{equation}
where $\lambda_{m} {=} 1$, $\lambda_{j} {=} 1$, $\lambda_{n} {=} 0.1$, $\lambda_{e} {=} 20$ in the experiments.

\begin{figure*}[tb]
\centering    
\includegraphics[width=\linewidth]{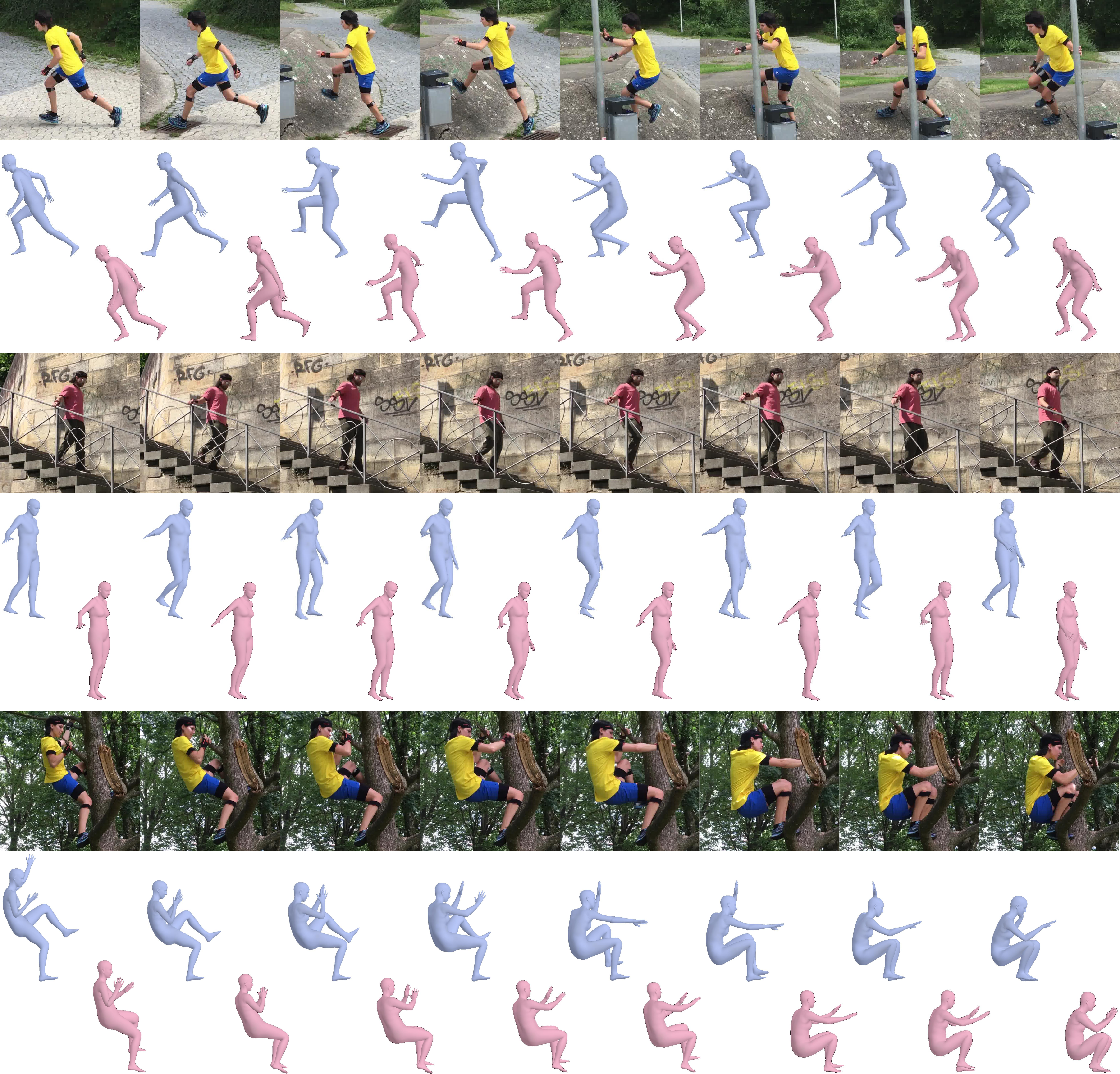}
\vspace{1mm}
\caption{Qualitative comparison between MPS-Net~\cite{wei2022capturing} and our PMCE. For each video sequence, the top rows show the video frames, the middle rows show the predicted mesh results from our PMCE~(blue), and the bottom rows show the mesh results from MPS-Net~(pink). Our method can produce more accurate and smooth 3D human motion in fast motions~(first sequence), occlusions~(second sequence), and slight pose changes~(last sequence).}
\label{fig:sup_comp}
\end{figure*}

\begin{figure*}[tb]
\centering    
\includegraphics[width=\linewidth]{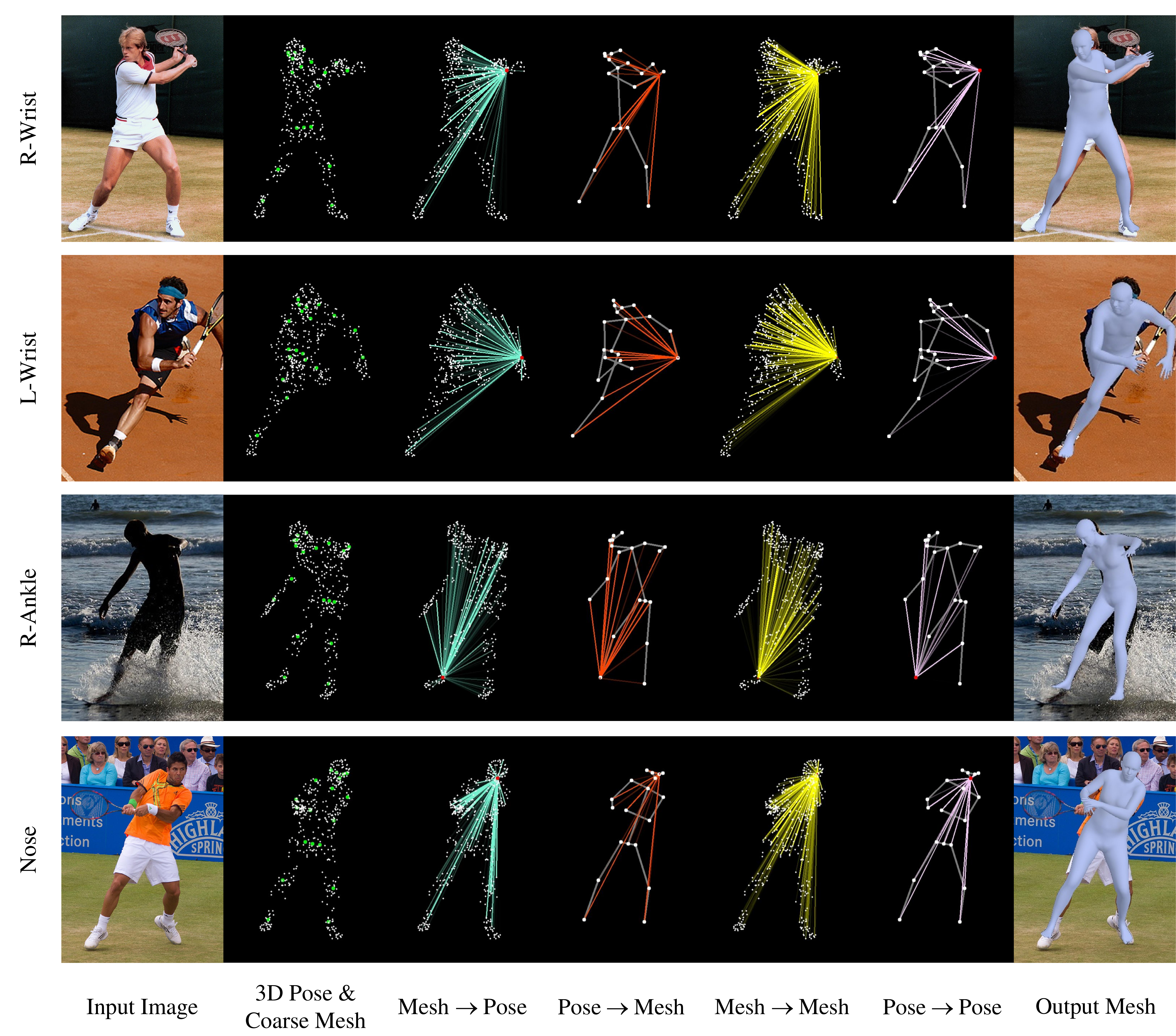}
\vspace{0.3mm}
\caption{Visualization of attention maps. From left to right: input image, the generated 3D pose and coarse mesh, four kinds of interactions, and the output mesh. `$\to$' denotes the direction of information flow. The brighter color indicates higher attention. And the color of lines in each attention map is normalized with the corresponding maximum. In Col.~3 `Mesh $\to$ Pose' interaction, the joint learns human body shape information from vertices. In Col.~4 `Pose $\to$ Mesh' interaction, the vertex can be guided by joints to perform mesh deformation. 
}
\label{fig:attn_map}
\end{figure*}

\vspace{4.5mm}
\noindent \textbf{\large{E. Additional Visualization Results}}
\vspace{2.3mm}

\noindent \textbf{Qualitative Comparison.}
Figure~\ref{fig:sup_comp} shows the qualitative comparison between the previous state-of-the-art video-based method MPS-Net~\cite{wei2022capturing} and our PMCE on the challenging video sequences. It shows that our method can produce more accurate and temporally consistent mesh results, especially in fast motions, occlusions, and delicate body deformations.

\noindent \textbf{Visualization of Attention Maps.}
We further study the interactions of pose and mesh in the proposed co-evolution decoder, including Mesh $\to$ Pose, Pose $\to$ Mesh, Mesh $\to$ Mesh, and Pose $\to$ Pose interactions. We obtain the above four kinds of attention maps from the last layer of the co-evolution decoder by averaging the attention values of all attention heads in their corresponding attention blocks. Figure~\ref{fig:attn_map} shows the visualization of attention maps taking different reference nodes. In `Mesh $\to$ Pose' interaction~(Col.~3), each joint can obtain the global shape information from vertices which is not available in its original pose representations. In `Pose $\to$ Mesh' interaction~(Col.~4), each mesh vertex aggregates pose information that can guide the mesh deformation. And in `Mesh $\to$ Mesh'~(Col.~5) and `Pose $\to$ Pose'~(Col.~6) interactions, vertices and joints perform internal adjustments, respectively. 

\end{document}